\documentclass[runningheads]{llncs}
\usepackage{graphicx}
\usepackage{amssymb}
\usepackage{amsmath}
\usepackage{graphicx}
\usepackage{wasysym}
\usepackage{booktabs}
\usepackage{makecell}
\usepackage{color}
\usepackage{subcaption}
\usepackage{wrapfig}

\usepackage[misc]{ifsym}
\usepackage{enumitem}
\setlist{nolistsep}

\usepackage{xcolor}

\usepackage{hyperref}
\hypersetup{
  colorlinks,
  citecolor=blue,
  linkcolor=red,
  urlcolor=blue}
 
\usepackage{multirow}
\usepackage{colortbl}
\definecolor{mygray}{RGB}{240,240,240}

\begin{document}
%
%\title{Med3DInsight: Exploring the Potential of Multi-modal Large Language Models for 3D Medical Understanding}
\title{Med3DInsight: Enhancing 3D Medical Image Understanding with 2D Multi-Modal Large Language Models}
\titlerunning{Med3DInsight}
\author{Qiuhui Chen
\and
Huping Ye
\and 
Yi Hong
\textsuperscript{(\Letter)}
}
% \author{***}
%
\authorrunning{Chen et al.}
% \authorrunning{***}
\institute{
Department of Computer Science and Engineering, \\ Shanghai Jiao Tong University, Shanghai, 200240, China \\
\email{yi.hong@sjtu.edu.cn}\\
}
% \institute{***}
%
\maketitle              % typeset the header of the contribution
\begin{abstract}
Understanding 3D medical image volumes is a critical task in the medical domain. However, existing 3D convolution and transformer-based methods have limited semantic understanding of an image volume and also need a large set of volumes for training. Recent advances in multi-modal large language models (MLLMs) provide a new and promising way to understand images with the help of text descriptions. However, most current MLLMs are designed for 2D natural images. To enhance the 3D medical image understanding with 2D MLLMs, we propose a novel pre-training framework called Med3DInsight, which marries existing 3D image encoders with 2D MLLMs and bridges them via a designed Plane-Slice-Aware Transformer (PSAT) module. Extensive experiments demonstrate our SOTA performance on two downstream segmentation and classification tasks, including three public datasets with CT and MRI modalities and comparison to more than ten baselines. Med3DInsight can be easily integrated into any current 3D medical image understanding network and improves its performance by a good margin. Our source code is publicly available at \href{https://github.com/Qybc/Med3DInsight}{https://github.com/Qybc/Med3DInsight}.

% The recognition capabilities of current state-of-the-art 3D models are limited by datasets with a small number of annotated data and a pre-defined set of categories. In its 2D counterpart, recent advances have shown that similar problems can be significantly alleviated by employing knowledge from other modalities, such as language. Inspired by this, leveraging multimodal information for 3D modality could be promising to improve 3D medical understanding under the restricted data regime, but this line of research is not well studied. 
% Therefore, we introduce Med3DInsight, which takes 3D medical data as input and harnesses the remarkable reasoning capabilities of 2D multi-modal large language models (2D MLLMs) to gain a comprehensive understanding of 3D medical scenes.
% To overcome the shortage of training triplets, Med3DInsight leverages a pre-trained 2D multi-modal large language model(2D MLLM), then learns a 3D representation space aligned with the common image-text space. 
% Med3DInsight is agnostic to 3D backbone networks and can easily be integrated into any 3D architecture. Experiments show that Med3DInsight effectively improves the performance of 3D backbones by simply pre-training them on multi 3D datasets using our framework, achieving state-of-the-art performance in both 3D classification and 3D segmentation on MM-WHS, CHAOS and OASIS. Our source code will be available online.

\end{abstract}
\section{Introduction}
%\fixme{First talk about Medical image understanding and its challenges, then talk about MLLM, then talk about our methods and contributions}

\begin{figure}[htb]
\includegraphics[width=\textwidth]{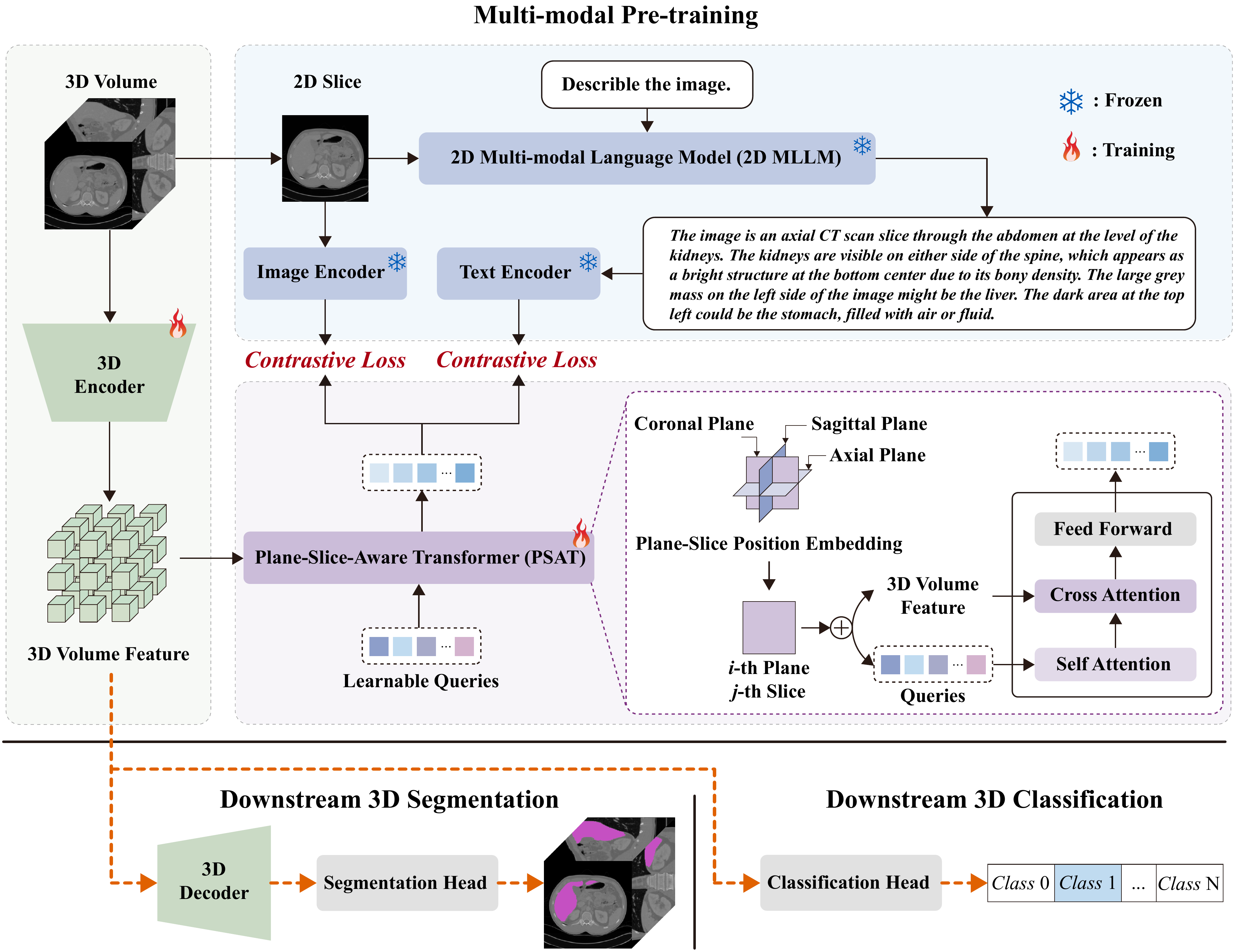}
\caption{Overview of our Med3DInsight framework. During pre-training (\textbf{Top}), we train a 3D image encoder and a plan-slice-aware transformer that aligns 3D image features with 2D slices and text features extracted from 2D MLLMs.
The pre-trained 3D encoder is further fine-tuned in the downstream tasks, including 3D segmentation (\textbf{Bottom Left}) and 3D classification (\textbf{Bottom Right}).
% The pertained 3D encoder is applied to downstream image segmentation and classification. 
% This framework aligns the 3D medical modality with the 2D vision-language embedding space, enabling us to leverage the MLLM for a comprehensive understanding of 3D medical scenes.
} 
\label{fig:overview}
\end{figure}

In medical research, analyzing and interpreting three-dimensional medical images is a critical task, which extracts valuable information for diagnosis, treatment planning, and other research in the field of healthcare. To understand 3D medical images, researchers often design different models according to specific purposes, e.g., 3D image classification models~\cite{jang2022m3t,bien2018deep}, 3D medical image segmentation models~\cite{chen2019med3d,hatamizadeh2022unetr,zhou2023nnformer}. Also, in the deep learning era, these models have limited semantic understanding of a 3D medical image, whose feature extraction is treated as a black box. We aim to design a pre-training framework for learning a general 3D medical image representation that has enhanced the semantic understanding of medical content in 3D scans, which can apply to multiple downstream tasks, including both image classification and segmentation.

Language description is an efficient way to improve image understanding through the integration of natural language processing and computer vision techniques. Recently, multi-modal large language models (MLLMs)~\cite{li2023blip,alayrac2022flamingo,2023GPT4VisionSC} have demonstrated impressive capabilities in enabling natural language to discuss and comprehend various visual scenes. Despite MLLMs excelling at processing 2D image content, their comprehension of the more challenging 3D medical volumes remains an open question. A couple of works explore the potential of leveraging existing MLLMs for 3D medical understanding, such as MedBLIP~\cite{chen2023medblip}, GTGM~\cite{chen2023generative}, and T3D~\cite{liu2023t3d}.  These methods either project 3D images to 2D representation as inputs of MLLMs~\cite{chen2023generative} or augment 2D MLLMs with 3D adaptors~\cite{chen2023medblip}, which are not pure 3D representation learners with losing the whole understanding of a 3D image. Alslthough these methods align image and text features, unlike the image encoder in an MLLM, they lack the semantic understanding of a 3D image. 

To address the above challenges, we propose to enhance a 3D image encoder using the image-text alignment technique as in MLLMs. However, unlike those in the computer vision domain, medical datasets with image-text pairs for 3D medical scans are extremely rare, even collecting a large set of 2D medical image-text pairs is a non-trivial task. 
In addition, there is a large gap between 3D medical image understanding with the current MLLMs designed for 2D nature images. 
Fortunately, GPT-4V(ision)~\cite{2023GPT4VisionSC} vision has already developed abilities for 2D medical imaging explanations~\cite{yang2023performance}.
Therefore, we introduce a novel 3D medical image representation learning framework, i.e., Med3DInsight, which aligns 3D image understanding with the help of 2D MLLMs, as shown in Fig.~\ref{fig:overview}. 

Mde3DInsight aligns 3D volume features with both 2D image and text features via contrastive learning~\cite{radford2021learning}. To bridge the gap between 3D and 2D feature spaces, we design a Plane-Slice-Aware Transformer (PSAT) module, which embeds the plane and slice position of a slice in a 3D volume and adopts learnable query techinque~\cite{carion2020end} to project 3D features into 2D feature space for mapping. 
We adopt GPT-4V(ision)~\cite{2023GPT4VisionSC} to generate detailed text descriptions for 2D medical slices that are extracted from 3D image scans.
To align image and text modalities in the same feature space~\cite{xue2023ulip}, we fine-tune image and text encoders of CLIP~\cite{radford2021learning} using the slice-text pairs generated by GPT-4V. 
% To obtain a large set of 2D medical image-text pairs for fine-tuning CLIP~\cite{radford2021learning} image and text encoders, w
During pre-training, we keep the GPT-4V and CLIP model frozen and train the 3D image encoder and PSAT module, resulting in a pretrained 3D image encoder for downstream segmentation and classification tasks.

Overall, our contributions are summarized as follows:
\begin{itemize}%[noitemsep, topsep=0pt]
    \item We propose a new framework Med3DInsight, which leverages 2D MLLMs to enhance medical image understanding of an existing 3D encoder. Our framework is general and improves multiple recent models, resulting in the state-of-the art (SOTA) performance for both 3D segmentation and classification on MM-WHS~\cite{zhuang2016multi}, CHAOS~\cite{kavur2021chaos}, and OASIS~\cite{marcus2007open} datasets.
    \item We design a Plane-Slice-Aware Transformer (PSAT) module that bridges the 3D medical image encoder with the 2D vision-language models. This module allows learning a mapping that is aware of the spatial orientation of visual features, which can be applied in other applications with need of establishing 2D and 3D feature mappings.  
\end{itemize}

\section{Methodology}
Fig.~\ref{fig:overview} presents the overview of our Med3DInsight framework, which can be divided into pre-training and downstream tasks. Pre-training includes three components, i.e., the 3D image encoder, the 2D MLLMs, and the Plane-Slice-Aware Transformer (PSAT) module. Regarding the 3D image encoder, we can choose an encoder of an existing model, such as SETR~\cite{zheng2021rethinking}, UNETR~\cite{hatamizadeh2022unetr}, nnFormer~\cite{zhou2023nnformer}, Vit~\cite{dosovitskiy2020image}, Swin-ViT~\cite{liu2022video}, etc.
% , which is the same case for the 3D decoder in the downstream segmentation tasks. 
Based on the image volumes used for pre-training, we first create slice and text pairs for 2D MLLMs and then use the PSAT module to make the connection between 2D and 3D feature spaces. 

%The initial column showcases our 3D medical image feature extractor, which processes the 3D medical image input to derive a 3D image feature.  
%The Plane-Slice-Aware Transformer (PSAT) accepts medical image embedding and learnable queries as input, with the output queries serving as input to the alignment of 2D slice image and text description. 
%In the PSAT, we introduce plane-slice position embeddings into the 3D medical feature along with corresponding queries to enhance the capability of spatial orientation representation. 

\noindent 
\textbf{Data Collection and 2D Feature Extraction.} We choose a public 3D medical dataset 3DSeg-8~\cite{chen2019med3d} for pre-training. This dataset has a collection of 2K images of various modalities, e.g., MRI, CT, and different body parts, including brain, heart, prostate, spleen, etc. These image scans are collected from multiple centers and have various spatial resolutions and intensity ranges. For each 3D medical volume image $k$ in this dataset, we create a triplet $D_k:(V_k, I_k^{(i,j)}, T_k^{(i,j)})$ of a medical volume $V_k$, an image slice $I_k^{(i,j)}$ extracted from the i-th plane (i.e., coronal, sagittal, or axial plane) and the j-th slice, and its corresponding text description $T_k^{(i,j)}$.  %Med3DInsight will then use these triplets for pre-training. 

%We build our dataset of triplets from 3DSeg-8~\cite{chen2019med3d}, which is one of the most extensive public 3D medical volume datasets. 3DSeg-8~\cite{chen2019med3d} is from different modalities (MRI and CT), distinctive scan regions (e.g. brain, heart, prostate, spleen, etc.) as well as multi-centers, which create a large variety of data characteristics, such as the spatial resolutions in 3D and range of pixel intensities. It contains around 2k 3D medical volume images, for each 3D medical volume image $i$ in the dataset, we create a triplet $T_i:(V_i, I_i,T_i)$ of medical volume $V_i$, slice image $I_i$ and text description $T_i$.  Med3DInsight will then use these triplets for pre-training.

%To obtain images that semantically align well with each 3D medical volume image, we synthesize multi-plane multi-slice of each volume image by random select each slice and extract the corresponding images from each plane.
%Specifically, we extract all slice images for every 3 plane(axial, coronal and sagittal). 

{\it (1) Slice and Feature Extraction.} To obtain the image slice set $\{I_k^{(i,j)}\}$, we extract all image slices of each image volume $V_k$ from all three planes. During each iteration of pre-training, we randomly select one image slice from a plane of an image volume as $I_k^{(i,j)}$ and take it as input of the image encoder $f_I (\cdot)$ adopted from CLIP to extract the its feature $h^I_{i,j,k} =  f_I (I_k^{(i,j)})$, as shown in Fig.~\ref{fig:overview}.
% \begin{align}
%     h^I_i = f_I (I_i)
% \end{align}

{\it (2) Text Generation and Feature Extraction.}
We employ a 2D MLLM (e.g., GPT-4V~\cite{2023GPT4VisionSC}) to generate the text descriptions $T_k^{(i,j)}$ of a specific 2D slice $I_k^{(i,j)}$, as shown in Fig.~\ref{fig:overview}. Then, we input this text description into our text encoder $f_T(\cdot)$ adopted from CLIP and obtain the text feature representation of the image slice $I_k^{(i,j)}$, i.e., $h^T_{i,j,k} =  f_T (I_k^{(i,j)})$.

In this way, we convert a triplet data $D_k:(V_k, I_k^{(i,j)}, T_k^{(i,j)})$ into a triplet feature set $H_k:(h^V_k, h^I_{(i,j,k)}, h^T_{(i,j,k)})$, where $h^V_k = f_V(V_k)$ and $f_V(\cdot)$ is the selected 3D image encoder.

% \begin{align}
%     h_i^T = f_T(T_i)
% \end{align}

% \subsubsection{Aligning Representations of Three Modalities}
% With the created triplets of volume, image, and text, Med3DInsight conducts pre-training to align representations of all three modalities into the same feature space. 

% Specifically, we take advantage of pre-trained vision-language models, i.e., CLIP, and train a 3D encoder by aligning the 3D feature with the features of image and text encoders ($f_I (\cdot)$ and $f_T(\cdot)$) of CLIP. 

% Our objective is to project these 3D volume features into the image and text embedding space of a pre-trained CLIP through our proposed Plane-Slice-Aware Transformer (PSAT).

\noindent
\textbf{PSAT Alignment.} In the feature sets $\{H_k\}$, we have both 3D image features $\{h^V_k\}$ and 2D image and text feature pairs $\{(h^I_{(i,j,k)}, h^T_{(i,j,k)})\}$, an alignment is desired to bridge the 2D and 3D gap and allow for the self-training based on contrastive learning. Here, we adopt the query techinque~\cite{carion2020end} augmented by knowing the plane-slice position of the randomly selected image slice. 

As shown in Fig.~\ref{fig:overview}, the PSAT module includes a set of $Q$ learnable query embeddings. In the experiments, we set $Q$ to 300 for a convenient projection into the image and text embedding space of the CLIP. 
% These queries interact with the volume feature $h^V_k$ through a cross-attention layer and produces an output $Q$ encoded visual vectors, one for each query embedding. 
These learnable queries interact with each other through a self-attention layer, then interact with the volume feature $h^V_k$ through a cross-attention layer and produces an output $Q$ encoded visual vectors, one for each query embedding.
These vectors then undergo processing through a multi-layer perceptron (MLP), resulting in the projection of the volume embedding, which is fed into the alignment with the image slice and text embedding via a contrastive loss~\cite{radford2021learning}.

Since the image slice $T_k^{(i,j)}$ is extracted from the image volume $V_k$ and located at the $j$-th slice of the $i$-th plane, the 3D-2D projection should be aware of this orientation relationship between image volume and slice. Therefore, we introduce a plane-slice position embedding for the volume feature to promote the model’s capacity of learning orientation and geometric relationships. 

Specifically, we first construct the plane-slice position embedding $V_p\in \mathbb{R}^{C \times I \times J}$ with zero initial parameters, where $C$ is the embedding dim of model, $I$ is the number of planes, and $J$ is the number of slices.
% Then we split the volume feature according to specific planes and slices. \fixme{this sentence is unclear.} 
During pre-training, when dealing with a slice image $I_k^{(i,j)}$, for instance, when a sample related to the eight slice (i.e., $i=8$) of the coronal plane (i.e., $j=0$), we only inject the eight slice at the coronal plane position embedding $V_p\in \mathbb{R}^{c \times 1 \times 1}$ into volume feature and queries, following general positional encoding~\cite{vaswani2017attention}.

% we inject the embedding corresponding position $(i,j)$ into both the volume feature and queries. \fixme{how?} For instance, when training a sample related to the eight slice (i.e., $i=8$) of the coronal plane (i.e., $j=0$), we only inject the eight slice at the coronal plane position embedding $V_p\in \mathbb{R}^{c \times 1}$ into volume feature and queries. \fixme{unclear, how to embed? give a reference? why 1 here?}

% Specifically, we first construct the plane-slice position embedding $V_p\in \mathbb{R}^{c \times 256}$ with zero initial parameters. \fixme{what is p? c?} Then we split the volume feature according to specific planes and slices. \fixme{this sentence is unclear.} During pre-training, when dealing with a slice image $I_k^{(i,j)}$, we inject the embedding corresponding position $(i,j)$ into both the volume feature and queries. \fixme{how?} For instance, when training a sample related to the eight slice (i.e., $i=8$) of the coronal plane (i.e., $j=0$), we only inject the eight slice at the coronal plane position embedding $V_p\in \mathbb{R}^{c \times 1}$ into volume feature and queries. \fixme{unclear, how to embed? give a reference? why 1 here?}

By aligning to 2D image slice and text features, we enhance the 3D medical image understanding in a self-learning way. Meanwhile, it has the potential to reduce the data requirement of image volumes, since the model is built upon pre-trained 2D MLLMs and we have millions of 2D image slices from the 2K image volumes of our pre-training dataset, which is demonstrated in our experiments.   

%By doing so, we hope that the abundant semantics already captured and aligned by CLIP’s encoders can be employed for better 3D understanding. The resulting unified feature space enables numerous cross-modal applications among these three modalities and potentially improves the 3D recognition performance of the underlying 3D backbone encoder $f_V(\cdot)$.

% \vspace{-0.2cm}
\section{Experiments}
To demonstrate the effectiveness of our pre-training 3D framework Med3DInsight, we conduct experiments on two groups of downstream tasks: 3D segmentation, including cardiac structure segmentation, abdominal organ segmentation, and brain segmentation, and 3D classification of Alzherimer's Disease (AD) and Normal Controls (NC). Then, we present detailed ablation studies and qualitative analysis to provide a deeper understanding of our approach.

%In this section, we first present downstream datasets, and implementation details. Then we present the quantitative results of standard 3D segmentation and 3D classification, respectively. Finally, we present detailed ablation studies and qualitative analysis to provide a deeper understanding of our approach.

\subsection{Datasets and Experimental Settings}
%\subsubsection{Cardiac Substructure Segmentation.}
\noindent
\textbf{MM-WHS~\cite{zhuang2016multi}.}
For cardiac segmentation, we used the Multi-Modality Whole Heart Segmentation (MM-WHS) Challenge 2017 dataset~\cite{zhuang2016multi}, consisting of unpaired 20 CT and 20 MRI scans with ground-truth pixel-level annotations acquired in the real clinical environment, including five labels, i.e., i.e., left ventricle blood cavity (LVC), right ventricle blood cavity (RVC), left atrium blood cavity (LAC), right atrium blood cavity (LAC), and ascending aorta (AA). In this experiment, we use CTs, consist of 177 to 363 slices with 512×512 pixels and voxel spacing ranging from 0.3 to 0.6 mm.
% \fixme{image size, resolution}. 

%\subsubsection{Abdominal Organ Segmentation}
\noindent
\textbf{CHAOS~\cite{kavur2021chaos}.} For abdominal organ segmentation, we use the training dataset comprised of 20 abdominal CT images from the ISBI 2019 CHAOS Challenge~\cite{kavur2021chaos}, including only one abdominal organ, i.e., the liver, for segmentation. Each CT scan consist of 81 to 266 slices with 512×512 pixels and voxel spacing ranging from 0.6 to 2.0 mm.
% \fixme{image size, resolution}.

%\subsubsection{Brain Substructure Segmentation}
\noindent
\textbf{OASIS~\cite{marcus2007open}.} For brain segmentation, we use the open-access OASIS1 dataset~\cite{marcus2007open}, including 414 subjects pre-processed with FreeSurfer~\cite{fischl2012freesurfer} and SAMSEG~\cite{puonti2016fast}. Our method is applied to segment four brain substructures, i.e., Cortex, Gray-Matter (GM), White-Matter (WM), and Cerebrospinal fluid (CSF). Each MRI scan's image size is 160×192×224 and voxel spacing is 1.0 mm.
For alzheimer's disease classification, we use the open-access OASIS2 dataset~\cite{marcus2010open}, consists of 335 T1-weighted sMRI scans collected from 135 subjects, including both AD subjects and healthy volunteers. Each MRI scan's image size is 224×224×224 and voxel spacing is 1.75 mm.

Since the image volumes used in the downstream datasets have various image resolutions and spacing, all image volumes are re-sampled into the isotropic voxel spacing of 1.0 mm in each dimension, and the image size of volumes ranges from 144×144×161 to 400×400×256. The voxel intensities of the images are then normalized to the range [0,1]. To simplify the preprocessing step, all images are first padded to a
cube shape and then scaled to a unified size of 128 × 128 × 128 as inputs. We {\it subject-wisely} split the data into training, validation, and testing with a ratio of 7:1:2.

% \fixme{classification datasets}

% \subsection{Implementation Details}
% Our Med3DInsight mainly comprises three components: 3D medical image feature extraction backbone, Plane-Slice-Aware Transformers (PSAT), and the 2D Multimodal Large Language Model (2D-MLLM). 

\noindent 
\textbf{Backbones, Baselines, and Other Settings.} 
For the 3D image encoder, we employ the SOTA 3D backbone, e.g., SETR~\cite{zheng2021rethinking}, UNETR~\cite{hatamizadeh2022unetr}, and nnFormer~\cite{zhou2023nnformer} for segmentation, ViT~\cite{dosovitskiy2020image} and Swin-ViT~\cite{liu2022video} for classification. For the segmentation task, we utilize both cross entropy loss and dice loss by simply averaging them. For the classification task, we utilize the cross-entropy loss.
All experiments are conducted on NVIDIA GeForce RTX 3090 GPUs.
For comparison, we select five baselines for 3D image segmentation, including UNet~\cite{ronneberger2015u}, AttUNet~\cite{oktay2018attention}, Med3D~\cite{chen2019med3d}, TransUNet~\cite{chen2021transunet}, nnUNet~\cite{isensee2021nnu}, and also five baselines for classification, including 3D ResNet50~\cite{he2016deep}, 3D DenseNet121~\cite{iandola2014densenet}, MRNet~\cite{bien2018deep}, MedicalNet~\cite{chen2019med3d}, and M3T~\cite{jang2022m3t}.  
For the PSAT module, we set the token number of learnable queries to 300, and the dimension of the token is 512. 
% In terms of the 2D-MLLM, we utilize GPT4V, which shows superior performance in the medical domain, and CLIP as our image-text encoders. We freeze the image and text encoders and only update the 3D encoder’s weights during pre-training.
Med3DInsight is trained for 300 epochs. We set the batch size as 32 and the learning rate as $1e-4$. AdamW is our optimizer for training Med3DInsight.

%For pretraining, we use contrastive loss following~\cite{radford2021learning}.

\begin{table}[t]
  \begin{center}
  % \tiny
    \caption{Comparison of 3D segmentation results on MM-WHS, CHAOS, and OASIS among our Med3DInsight and baseline methods. 
    %Med3DInsight significantly improves our baselines. 
    % Our best result outperforms SOTA largely by around 2\% on CLass-mean Dice.
    }
    \label{tab:results_seg}
    \begin{tabular}{ccccccccccccc} 
      \cmidrule(r){1-13}
      \multirow{2}*{Methods} & \multicolumn{6}{c}{\textbf{MM-WHS}} & \multicolumn{1}{c}{\textbf{CHAOS}}  & \multicolumn{5}{c}{\textbf{OASIS}} \\
      \cmidrule(r){2-7} \cmidrule(r){8-8} \cmidrule(r){9-13}
       & LVC & RVC& LAC & RAC & AA & Avg. & Liver & Cortex & GM & WM & CSF & Avg.\\
      \cmidrule(r){1-13}
      UNet~\cite{ronneberger2015u} & 80.9 & 78.1 & 76.6 & 72.3 & 74.7 & 76.5 & 79.9 & 70.9 & 83.3 & 83.2 & 80.5 & 79.5\\
      AttUNet~\cite{oktay2018attention} & 81.0 & 78.9 & 77.1 & 74.2 & 75.2 & 77.3 & 79.7 & 70.8 & 83.1 & 83.1 & 79.8 & 79.2\\
      Med3D~\cite{chen2019med3d} & 83.0 & 77.1 & 81.6 & 74.9 & 77.9 & 78.9 & 81.2 & 76.4 & 85.5 & 87.7 & 84.5 & 83.5\\
      TransUNet~\cite{chen2021transunet} & 82.9 & 76.9 & 80.8 & 76.1 & 78.6 & 79.1 & 82.6 & 76.6 & 85.6 & 87.3 & 84.7 & 83.6 \\
      nnUNet~\cite{isensee2021nnu} & 83.9 & 76.6 & 78.9 & 76.8 & 81.9 & 79.6 & 84.8 & 77.2 & 86.0 & 88.0 & 85.5 & 84.2\\
      
      \cmidrule(r){1-13}
      SETR~\cite{zheng2021rethinking} & 85.1 & 77.2 & 80.5 & 75.7 & 81.7 & 80.0 & 86.2 & 81.5 & 86.9 & 84.5 & 87.2 & 85.0 \\
      \rowcolor{mygray}
      +Med3DInsight & 85.2 & 81.5 & 81.6 & 80.4 & 85.7 & 82.9 & 89.8 & 82.6 & 89.1 & 88.3 & 90.1 & 87.5\\
        & & &  &  &  & \textcolor[RGB]{0,153,62}{+2.9}& \textcolor[RGB]{0,153,62}{+3.6} &  &  &  &  & \textcolor[RGB]{0,153,62}{+2.5}\\
      \cmidrule(r){1-13}
      UNetr~\cite{hatamizadeh2022unetr}  & 85.9 & 81.0 & 83.4 & 80.2 & 87.2 & 83.5 & 89.9 & 80.3 & 88.0 & 89.5 & 88.2 & 86.5 \\
      \rowcolor{mygray}
      +Med3DInsight  & 86.5 & 81.2 & 85.3 & 83.5 & 89.7 & 85.2 & 91.2 & 83.0 & 90.1 & 91.3 & 90.4 & 88.7 \\
       & & & & & & \textcolor[RGB]{0,153,62}{+1.7} & \textcolor[RGB]{0,153,62}{+1.3} & & & & &\textcolor[RGB]{0,153,62}{+2.2} \\
      \cmidrule(r){1-13}
      nnFormer~\cite{zhou2023nnformer} & 85.2 & 83.3 & 85.5 & 85.0 & \textbf{90.6} & 85.9 & 91.9 & 89.1 & 93.1 & 94.6 & 92.8 & 92.4 \\
      \rowcolor{mygray}
      +Med3DInsight & \textbf{89.3} & \textbf{89.1} & \textbf{86.5} & \textbf{87.4} & 90.5 & \textbf{88.6} & \textbf{94.1} & \textbf{92.6} & \textbf{94.2} & \textbf{96.8} & \textbf{95.3} & \textbf{94.7}\\
      & & & & & & \textcolor[RGB]{0,153,62}{+2.7} & \textcolor[RGB]{0,153,62}{+2.2} & & & & & \textcolor[RGB]{0,153,62}{+2.3}\\
      \cmidrule(r){1-13}
    \end{tabular}
  \end{center}
\end{table}

\begin{figure}[htb]
\begin{center}
\includegraphics[width=\textwidth]{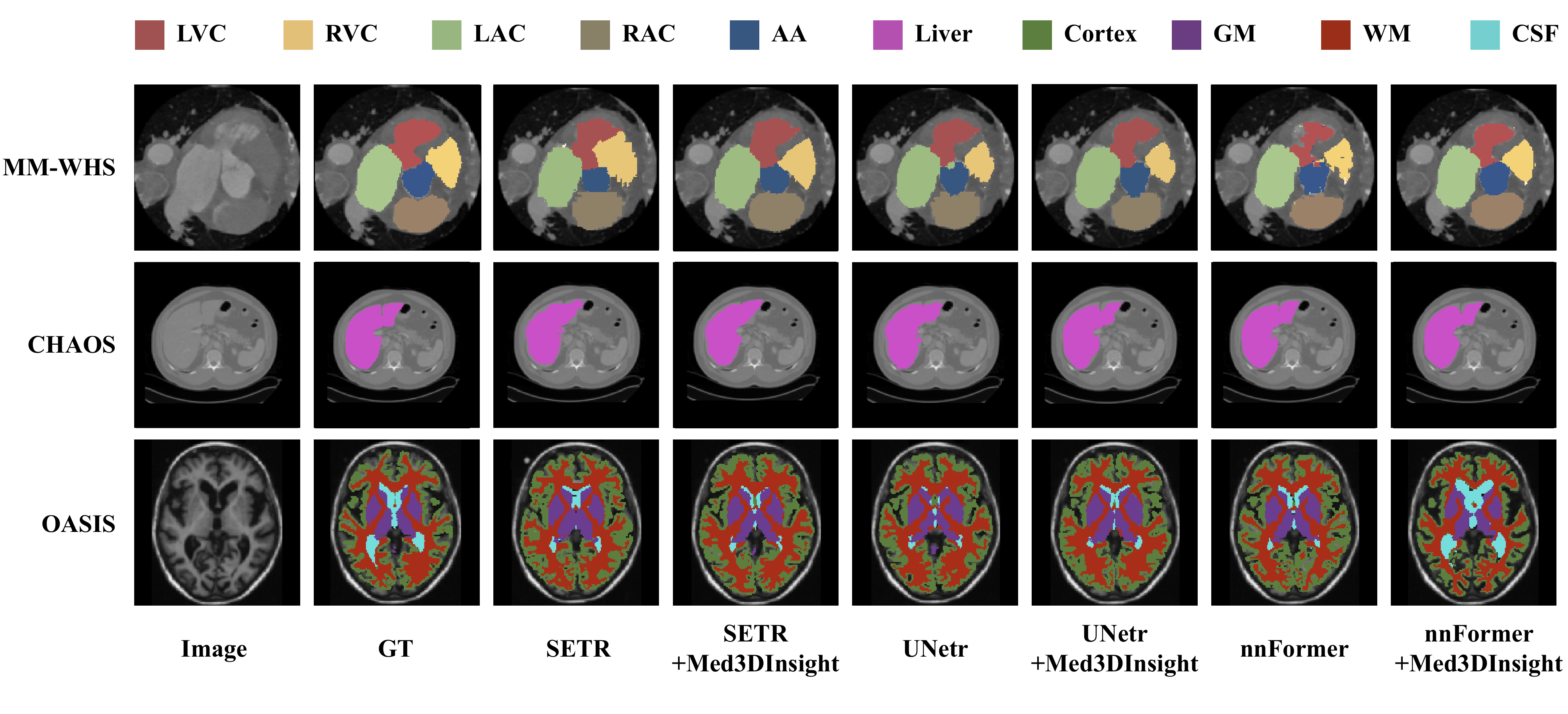}
\caption{Visual comparison of 3D segmentation results predicted by multiple methods and enhanced by our Med3DInsight.} 
\label{fig:vis}
\end{center}
\end{figure}

%\begin{table}[htp]
\begin{wraptable}{l}{6.7cm}
\vspace{-0.5cm}
  \begin{center}
  % \tiny
    \caption{Comparison of our Med3DInsight and baselines on OASIS 3D Classification. 
    %Med3DInsight significantly improves our baselines.
    }
    \label{tab:results_cls}
    \begin{tabular}{ccc} 
      \cmidrule(r){1-3}
      \multirow{2}*{Methods}   & \multicolumn{2}{c}{OASIS} \\
      \cmidrule(r){2-3}
       & Accuracy(\%) & AUC(\%)\\
      \cmidrule(r){1-3}
      3D ResNet50~\cite{he2016deep} & 66.2 & 65.9\\
      3D DenseNet121~\cite{iandola2014densenet} & 72.3 & 72.7 \\
      MRNet~\cite{bien2018deep} & 70.8 & 71.9\\
      MedicalNet~\cite{chen2019med3d} & 73.9 & 72.7 \\
      M3T~\cite{jang2022m3t} & 80.2 & 81.7 \\
      \cmidrule(r){1-3}
      ViT~\cite{dosovitskiy2020image} & 80.5 & 82.6 \\
      \rowcolor{mygray}
      +Med3DInsight & 81.4 & 84.3 \\ 
      & \textcolor[RGB]{0,153,62}{+0.9} & \textcolor[RGB]{0,153,62}{+1.7} \\
      \cmidrule(r){1-3}
      Swin-ViT~\cite{liu2022video} & 82.8 & 84.4 \\
      \rowcolor{mygray}
      +Med3DInsight & \textbf{84.1} & \textbf{85.7} \\ 
      & \textcolor[RGB]{0,153,62}{+1.3} & \textcolor[RGB]{0,153,62}{+1.3} \\
      \cmidrule(r){1-3}
    \end{tabular}
  \end{center}
%\end{table}
\vspace{-0.5cm}
\end{wraptable}

\subsection{Experimental Results And Ablation Study}
Table~\ref{tab:results_seg} reports the 3D segmentation results on the MM-WHS, CHAOS, and OASIS datasets, and Table~\ref{tab:results_cls} reports the 3D classification results on OASIS. Experimental results show that Med3DInsight enhances all backbones and improves both segmentation and classification results by over 2\% mean Dice and 1\% classification accuracy, respectively. Also, all Med3DInsight-augmented methods outperform all baselines. For 3D segmentation, the combination of Med3DInsight with nnFormer achieves the SOTA performance on all three segmentation tasks. For 3D classification, the combination of Swin-ViT with Med3DInsight performs the best on the OASIS dataset.
Fig.~\ref{fig:vis} visualizes the qualitative comparison of 3D segmentation results. Med3DInsight shows improved segmentation performance of baselines. Specifically, our model demonstrates better performance in capturing the fine-grained details of organs or substructures.

% %\begin{table}[htp]
% \begin{wraptable}{r}{0.55\textwidth}
%   \begin{center}
%   % \tiny
%     \caption{Ablation study of PSAT. QTrans denotes the Query Transformer, and PSPE denotes the Plane-Slice Position Embedding.
%     }
%     \label{tab:ab1}
%     \begin{tabular}{c|cc|ccccc} 
%       \cmidrule(r){1-8}
%       & QTrans & PSPE & Cortex & GM & WM & CSF & Avg. \\
%       \cmidrule(r){1-8}
%         $Ex1$ & & & 88.3 & 92.6 & 95.1 & 93.8 & 92.5 \\
%         $Ex2$ & \checkmark & & 91.7 & 94.3 & 95.2 & 94.5 & 93.9  \\ 
%         $Ex3$ & \checkmark & \checkmark & 92.6 & 94.2 & 96.8 & 95.3 & 94.7  \\
%       \cmidrule(r){1-8}
%     \end{tabular}
%   \end{center}
% \end{wraptable}
% %\end{table}

\begin{wrapfigure}{l}{0.6\textwidth}
\begin{minipage}{\linewidth}
\begin{center}
    \captionof{table}{Ablation study of PSAT. QTrans denotes the Query Transformer, and PSPE denotes the Plane-Slice Position Embedding.
    }
    \label{tab:ab1}
    \begin{tabular}{c|cc|ccccc} 
      \cmidrule(r){1-8}
      & QTrans & PSPE & Cortex & GM & WM & CSF & Avg. \\
      \cmidrule(r){1-8}
        $Ex1$ & & & 88.3 & 92.6 & 95.1 & 93.8 & 92.5 \\
        $Ex2$ & \checkmark & & 91.7 & 94.3 & 95.2 & 94.5 & 93.9  \\ 
        $Ex3$ & \checkmark & \checkmark & 92.6 & 94.2 & 96.8 & 95.3 & 94.7  \\
      \cmidrule(r){1-8}
    \end{tabular}
  \end{center}
\end{minipage}  

\vspace{0.1cm}

\begin{minipage}{\linewidth}
        \centering
		\includegraphics[width=0.99\columnwidth]{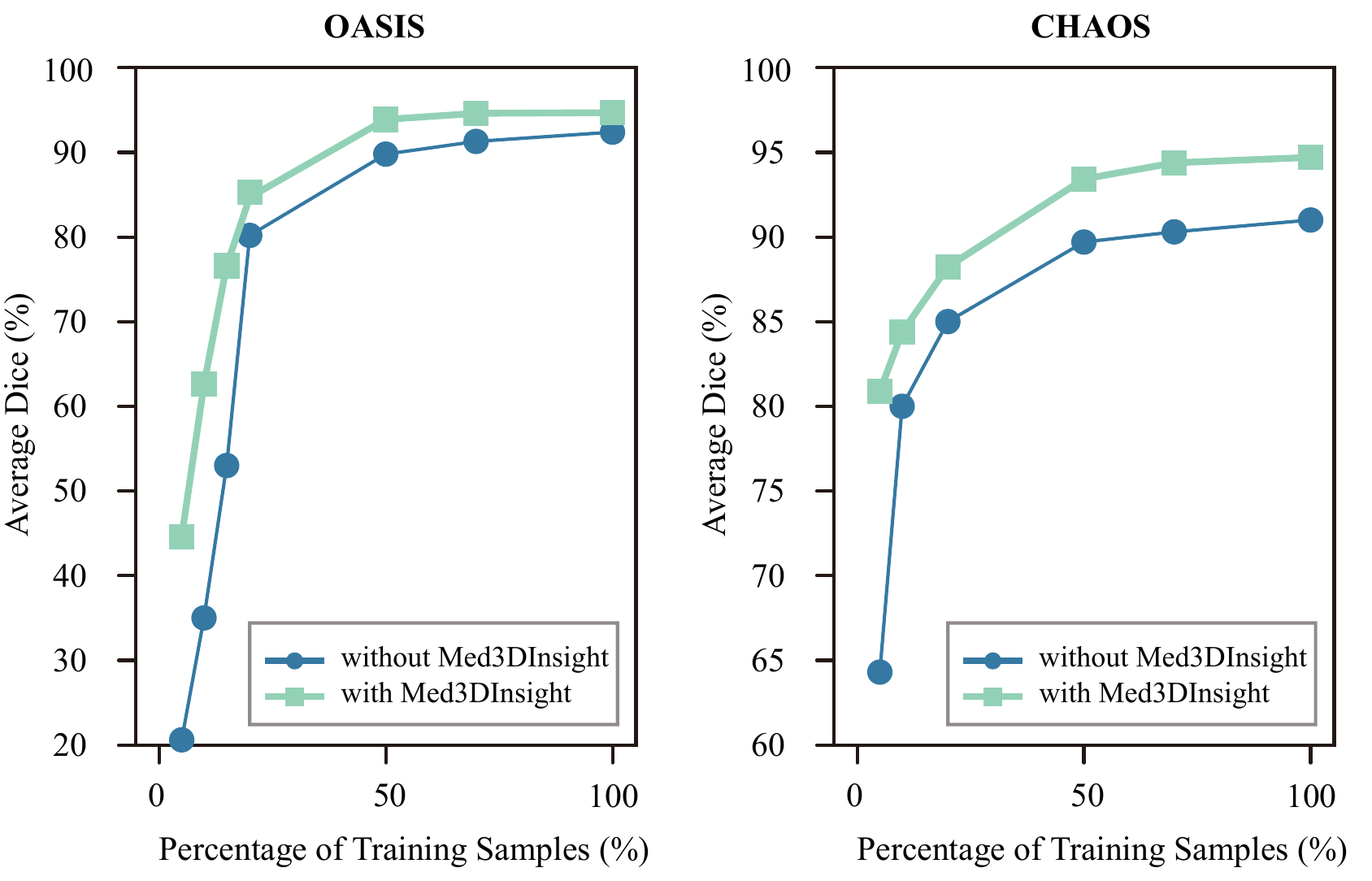}
		\caption{Ablation study of data efficiency with and without our Med3DInsight.}
		\label{fig:ab1}
\end{minipage}
\vspace{-0.5cm}
% \end{figure}
\end{wrapfigure}

%\subsection{Ablation Study}
\noindent
\textbf{Effectiveness of the PSAT Model.}
To demonstrate the effectiveness of each component in PSAT, we perform an ablation study on the OASIS 3D segmentation task. 
As shown in Table~\ref{tab:ab1}, $Ex1$ denotes the absence of any transformer structure and position embedding. In this case, the 3D image feature is directly input to a projection layer to align the dimensions and is then fed to the calculate contrastive loss to align with the features of 2D vision-language of CLIP.
In $Ex2$, with the use of Query Transformer, we observe that the dice score achieves 93.9\%. The results demonstrate that the structure of the Query Transformer can better align 3D medical image features with 2D vision-language features.
Compared with $Ex2$ and $Ex3$, the introduction of the Plane-Slice Position embedding achieves an 0.8\% dice score improvement.
This set of results affirms that incorporating the Plane-Slice Position embedding aids the model in better understanding 3D scenes and spatial relationships.

% %\begin{figure}[htb]
% \begin{wrapfigure}{r}{0.55\textwidth}
% \vspace{-0.4cm}
% \begin{center}
% \includegraphics[width=0.52\textwidth]{ablation.pdf}
% \caption{Ablation study of data efficiency with and without our Med3DInsight.} 
% %The X axis indicates the percentage of samples used for training and Y axis denotes the average dice.} 
% \label{fig:ab1}
% \end{center}
% \end{wrapfigure}
% %\end{figure}

% \begin{figure}[t]

\noindent
\textbf{Data Efficiency.}
Model pre-training can potentially reduce the demand for labeled data in downstream tasks. We validate the data efficiency by comparing the effectiveness of using our Med3DInsight based on nnFormer, under a varying number of fine-tuning samples. Fig.~\ref{fig:ab1} presents the comparison results, showing that  
the nnFormer's performance is largely improved in the low data regime when being pre-trained under the Med3DInsight framework. 
We observe that Med3DInsight performs better on CHAOS than on the OASIS dataset when using less than 20\% training data. 
This is probably because our model is pre-trained on the 3DSeg-8 dataset, which has a large amount of abdominal CT images. Even in this situation, Med3DInsight still improves the performance on OASIS by a good margin.

\section{Conclusion and Discussion}
In this paper, we propose a new pre-training framework Med3DInsight, which utilizes 2D MLLMs to enhance medical image understanding and improve the downstream segmentation and classification performance of multiple current 3D image understanding networks. To bridge the feature space gap between the 3D image encoder and 2D MLLMs, we introduce the Plane-Slice-Aware Transformer (PSAT) based on the learnable query technique. Experimental results demonstrate the consistent improvements of our method in downstream tasks. In future work, we consider further exploring the semantic understanding of the 3D image volume in our framework, by integrating with large language models for 3D image caption or 3D visual question answering.

\bibliographystyle{splncs04}
\bibliography{refs}

\end{document}